\pdfoutput=1

\documentclass[11pt]{article}

\usepackage{coling}

\usepackage{times}
\usepackage{latexsym}

\usepackage[T1]{fontenc}

\usepackage[utf8]{inputenc}

\usepackage{microtype}

\usepackage{inconsolata}

\usepackage{graphicx}

\usepackage{microtype}
\usepackage{import}
\usepackage{layout}
\usepackage{tabularx, makecell}
\usepackage{booktabs}
\usepackage{mathrsfs}
\usepackage{amssymb} 
\usepackage{url}
\usepackage{hyperref}
\usepackage{graphicx}
\usepackage{xspace,paralist}
\usepackage{times,latexsym}
\usepackage{amsmath}
\usepackage{appendix}
\usepackage{comment} 
\usepackage{enumitem}
\usepackage{makecell}
\usepackage{multirow}
\usepackage{xcolor}
\usepackage{arydshln}
\usepackage{cleveref}
\usepackage{tcolorbox}
\usepackage{todonotes}
\usepackage{longtable,supertabular}
\usepackage{amssymb}
\usepackage{pifont}

\usepackage[frozencache,cachedir=.]{minted}
\definecolor{bg}{rgb}{0.95,0.95,0.95}

\newcommand{\cmark}{\textcolor{green}{\ding{51}}}
\newcommand{\xmark}{\textcolor{red}{\ding{55}}}

\newcommand{\ex}[1]{\textit{#1}\xspace}

\newcommand{\tabref}[2][]{Table#1~\ref{#2}\xspace}
\newcommand{\figref}[1]{Figure~\ref{#1}\xspace}

\newcommand{\appref}[1]{Appendix~\ref{#1}\xspace}

\newcommand{\checker}[1]{\textit{#1}\xspace}
\newcommand{\rarr}{\checker{RARR}}
\newcommand{\factscore}{\checker{FActScore}}
\newcommand{\factool}{\checker{FacTool}}
\newcommand{\factcheckgpt}{\checker{Factcheck-GPT}}
\newcommand{\cove}{\checker{CoVe}}
\newcommand{\longformsafe}{\checker{Longform SAFE}}

\newcommand{\model}[1]{\textit{#1}\xspace}
\newcommand{\perplexityai}{\model{Perplexity.ai}}
\newcommand{\gptfour}{\model{GPT-4}}
\newcommand{\gptfouro}{\model{GPT-4o}}
\newcommand{\llama}{\model{LLaMA}}

\newcommand{\dataset}[1]{\text{#1}\xspace}
\newcommand{\factoolqa}{\dataset{FacTool-QA}}
\newcommand{\felmwk}{\dataset{FELM-WK}}
\newcommand{\factcheckbench}{\dataset{Factcheck-Bench}}

\newcommand{\demo}[1]{\textsc{#1}\xspace}
\newcommand{\ofc}{\demo{OpenFactCheck}}
\newcommand{\loki}{\demo{Loki}}

\title{Loki: An Open-Source Tool for Fact Verification}

\author{Haonan Li$^{1,2}$ \quad Xudong Han$^{1,2}$ \quad Hao Wang$^{1}$,\\
\bf{Yuxia Wang$^{1,2}$ \quad Minghan Wang$^{3}$ \quad Rui Xing$^{2,4}$ \quad Yilin Geng$^{1,4}$}\\
\bf{Zenan Zhai$^{1}$ \quad Preslav Nakov$^{2}$ \quad Timothy Baldwin$^{1,2,4}$} \\
$^1$LibrAI \quad $^2$MBZUAI \quad $^3$Monash University  \quad $^4$The University of Melbourne \\
}

\begin{document}
\maketitle
\begin{abstract}
We introduce \loki, an open-source tool designed to address the growing problem of misinformation.
\loki adopts a human-centered approach, striking a balance between the quality of fact-checking and the cost of human involvement.
It decomposes the fact-checking task into a five-step pipeline: breaking down long texts into individual claims, assessing their check-worthiness, generating queries, retrieving evidence, and verifying the claims.
Instead of fully automating the claim verification process, \loki provides essential information at each step to assist human judgment, especially for general users such as journalists and content moderators. Moreover, it has been optimized for latency, robustness, and cost efficiency at a commercially usable level. 
\loki is released under an MIT license and is available on GitHub.\footnote{\url{https://github.com/Libr-AI/OpenFactVerification}} 
We also provide a video presenting the system and its capabilities.\footnote{\url{https://www.youtube.com/watch?v=L_3Dp41Lk_k}}
\end{abstract}

\section{Introduction}

In today's digital landscape, the rapid spread of misinformation has become a significant societal problem, with far-reaching consequences for politics, public health, and social stability \citep{pan2023riskmisinformationpollutionlarge, augenstein2023factualitychallengeseralarge}. With the rise of online platforms, users are exposed to large volumes of information, often without the ability to assess its accuracy. While manual fact-checking is reliable, it is labor-intensive, time-consuming, and often requires domain expertise, creating a gap where misinformation can spread unchecked and cause harm before being addressed.

To address this problem, automated fact-checking systems have been proposed, but have mostly focused on full automation, which can negatively impact quality. 

Here, we propose \loki, which offers a semi-automated, human-in-the-loop approach to fact verification. Instead of completely eliminating human participation, \loki assists users by breaking down the fact-checking process into five manageable steps, ensuring that human judgment remains integral to decision-making. This benefits users, such as journalists and content moderators, who need reliable tools to quickly and accurately verify information.

\loki offers a five-step pipeline for fact verification: identifying claims, assessing their check-worthiness, generating queries for evidence retrieval, retrieving evidence, and verifying the claims. This modular framework ensures flexibility and adaptability. It supports fact-checking in multiple languages and integration with large language models (LLMs). Additionally, \loki is optimized for practical use, with improvements in latency, robustness, and cost-efficiency, making it well-suited for commercial applications.

\loki is implemented in Python and offers ease of use through multiple interfaces: a user-friendly graphical interface, command-line functionality, and the ability to be imported as a package into other projects. 
To demonstrate its utility, we compare \loki against several recent fact-checking tools, and show that optimized system implementation and prompt engineering yields demonstrable improvements over other tools in terms of efficiency while achieving very competitive performance.


\begin{table*}[th]
    \centering
    \resizebox{\textwidth}{!}{
    \begin{tabular}{lccccccc}
    \toprule
    Fact-checking System & UI & Asynchronous & Multilingual & Multi-LLM & Granularity & Transparency & Open-source\\
    \midrule
    \rarr~\citep{gao2022attributed} & \xmark & \xmark & \xmark & \xmark & Document & \xmark & \cmark\\
    \factscore~\citep{min2023factscore} & \xmark & \xmark & \xmark & \xmark & Claim & \xmark & \cmark\\
    \factool~\citep{chern2023factool} & \xmark & \cmark & \xmark & \xmark & Claim & \xmark & \cmark\\
    \factcheckgpt~\cite{wang2023factcheck} & \xmark & \xmark & \xmark & \xmark & Claim & \cmark & \cmark\\
    \cove~\citep{dhuliawala2023chain} & \xmark & \xmark & \xmark & \xmark & Claim & \xmark & \cmark\\
    \longformsafe~\cite{wei2024longformfactuality} & \xmark & \xmark & \xmark & \cmark & Claim & \xmark & \cmark\\ 
    \perplexityai & \cmark & \cmark & \cmark & unclear & Claim & \xmark & \xmark \\
    \ofc~\cite{wang2024openfactcheck} & \cmark & \cmark & \xmark & \xmark & Claim & \cmark & \cmark\\
    \midrule
    \loki (ours) & \cmark & \cmark & \cmark & \cmark & Claim & \cmark & \cmark\\
    \bottomrule
    \end{tabular}
    }
    \caption{Comparison of representative automatic fact-checking \model{pipelines}, \demo{demos} and products in the last two years from seven perspectives: (1) \textbf{UI} --- the system has user interface supporting easy interaction with general users; (2) \textbf{Asynchronous} processing for retrieving evidence from web pages and calling LLM APIs; (3) \textbf{Multilingual} --- the system is designed to support languages other than English;
      (4) \textbf{Multi-LLM} --- flexibly calling different LLM APIs as fact verifiers; (5) \textbf{Granularity} --- the smallest granularity of document decomposition and verification supported by the system; (6) \textbf{Transparency} --- the system can show fine-grained snippets of evidence with the corresponding URL, and the relationship between the evidence and the claim (support, refute, or irrelevant); and
    (7) \textbf{Open-source} --- the system is open-sourced.}
    \label{tab:relatedwork}
\end{table*}

\section{Related Work}
Numerous automated fact-checking systems have been developed, including \rarr, \factscore, \factool, \factcheckgpt, and \longformsafe~\cite{gao2022attributed, min2023factscore, chern2023factool, wang2023factcheck, wei2024longformsafe, fadeeva-etal-2024-fact}. However, these tools are often inaccessible to general users who may not have a Python environment to compile code and run verification processes. Although these systems can serve as backends for various services, they lack a user-friendly web interface that allows users to verify text inputs by simply typing or pasting text and clicking a \texttt{check} button. \loki addresses this gap by providing an accessible, human-friendly user interface.

Each fact-checking system also has its own strengths. For instance, \factcheckgpt offers a fine-grained framework encompassing all possible subtasks to enhance the fact-checking process. \factool uses a low-latency evidence retriever through asynchronous processing, while \factscore introduces a scoring metric that calculates the percentage of true claims within a text, providing a quantitative measure of the input’s credibility. \loki integrates these advantages into a unified system~\cite{wang2024factuality-survey}.
\tabref{tab:relatedwork} compares eight representative fact-checking pipelines, demos, or products (e.g., \perplexityai) with \loki across seven dimensions. Of the systems surveyed, \loki is the only one to support all seven listed system aspects.


\section{Loki Fact Checker}

In this section, we provide an in-depth overview of \loki, focusing on its core functionalities and architectural design. We first outline the five-component fact verification pipeline and then discuss the architecture’s flexibility that allows component substitution for different domains. After that, we explore how parallelism is employed to enhance the system’s efficiency and reduce latency. Finally, we present the \loki user interface, designed to provide a seamless and user-friendly user experience.

\subsection{Fact Verification Pipeline}


Previous work structures fact verification into a series of steps, commonly including text decomposition, checkworthiness identification, evidence retrieval and collection, stance detection, and correction determination \citep{wang2024factcheckbenchfinegrainedevaluationbenchmark}. In this work, we propose a five-step pipeline consisting of the following modules: Decomposer, Checkworthiness Identifier, Query Generator, Evidence Retriever, and Claim Verifier.

\paragraph{Decomposer} breaks down long texts into smaller, atomic claims. It produces individual claims that can be verified independently. For a better user experience, we ensure the decomposed claims are traceable to the original text. During the result presentation, \loki displays both the original and decomposed claims for contextual clarity.

\paragraph{Checkworthiness Identifier} filters out unworthy claims that are vague, ambiguous, or opinion-based, ensuring only factual statements proceed for verification. For instance, claims like \ex{MBZUAI has a vast campus} are deemed unworthy due to the subjective interpretation of \ex{vast}. 

\paragraph{Query Generator} converts check-worthy claims into optimized queries for evidence retrieval, focusing on keyword-based retrieval.

\paragraph{Evidence Retriever} gathers relevant information to support the verification process. Currently, \loki retrieves evidence from online sources via search engine APIs. 

\paragraph{Claim Verifier} evaluates retrieved evidence to verify the claim, presenting supporting or refuting snippets for users to make informed judgments.

\subsection{Implementation and Extensibility}

\loki is implemented in Python and can be easily integrated into other projects as an importable package. While the core functionality leverages the capabilities of LLMs, we also provide implementations using traditional NLP tools like NLTK or SpaCy.\footnote{Our evaluation shows that LLM-based implementations are generally more robust and accurate in practical scenarios.} This section introduces \loki’s core implementation and its extensibility.

\paragraph{LLM-based Implementation} \loki employs LLMs in four of its five components: Decomposer, Checkworthiness Identifier, Query Generator, and Claim Verifier. Each component is implemented as a Python class with functions that interact with LLMs to perform core tasks. For instance, the Decomposer class includes functions to break down long texts into claims and map these claims back to the original text.

Each LLM-based function comes with a set of default prompts, typically including: (1) a brief task description, (2) input and output formats, and (3) few-shot examples to guide the LLM. To generate consistently parsable outputs, we prompt the LLMs to return results in a structured format, such as JSON.\footnote{Which is currently only well-supported in some models such as GPT4 \cite{openai2024gpt4technicalreport}.} This approach streamlines the integration of LLM outputs into subsequent steps of the pipeline. For each task, we hand-crafted 10 test cases and optimized the prompts to maximize LLM performance. An example prompt for the Decomposer is shown in \figref{fig:prompt_example}.

To maintain system reliability, all LLM function calls are wrapped with a retry mechanism that automatically retries in case of network issues or transient errors.


\begin{figure}[t]
    \scriptsize
    \centering
\begin{tcolorbox}[colframe=white, left=3mm, right=3mm]
Your task is to \textcolor{orange}{decompose the text into atomic claims.}\\
The answer should be a \textcolor{red}{JSON} with a single key ``claims'', with the value of a list of strings, where each string should be a context-independent claim, representing one fact.\\
Note that:\\
1. Each claim should be concise (less than 15 words) and self-contained.\\
2. Avoid vague references like `he', `she', `it', `this', `the company', `the man' and using complete names.\\
3. Generate at least one claim for each single sentence in the texts.\\

For example,\\
\textcolor{blue}{
Text: Mary is a five-year old girl, she likes playing piano and she doesn't like cookies.\\
Output:\\
\{\{``claims'': [``Mary is a five-year old girl.'', ``Mary likes playing piano.'', ``Mary doesn't like cookies.'']\}\}\\
}\\
Text: \{doc\}\\
Output:
\end{tcolorbox}
\caption{An example of the prompt for the Decomposer. This is an template and the actual prompt is generated by replacing the placeholder \{doc\} with the input text.}
    \label{fig:prompt_example}
\end{figure}

\begin{figure*}[t]
    \centering
    \includegraphics[width=.9\linewidth]{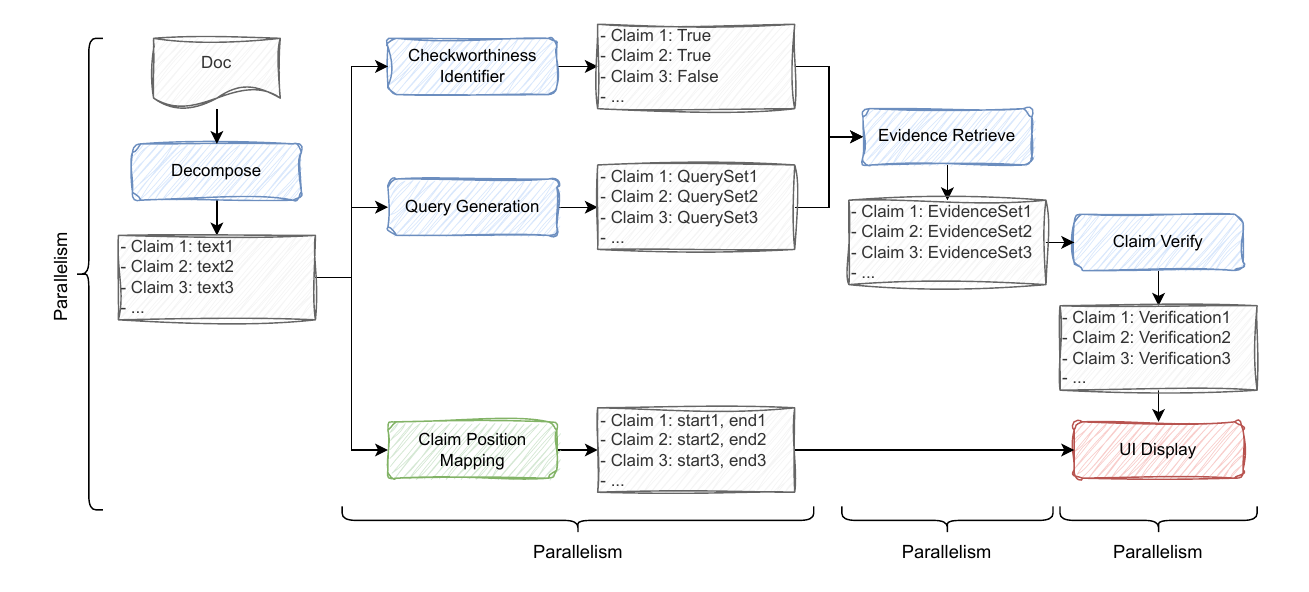}
    \caption{Parallel execution of independent components in the fact-checking pipeline. The entire process can be completed within the combined time of three LLM calls and one web query.}
    \label{fig:parallelism}
\end{figure*}

\paragraph{Evidence Retriever} 
Currently, \loki uses the Google Search API (Serper API) for open-domain questions, retrieving web-based evidence to support or refute claims.\footnote{\url{https://serper.dev/}} This module’s design is flexible,  allowing integration with alternative search engines or specialized databases to expand evidence sources. For each claim, \loki generates three distinct search queries.\footnote{Using the claim itself as a query can lead to biased or irrelevant results, as false claims may mislead the search.} If the Google Search API returns a direct answer, it is captured as primary evidence. Otherwise, the top five search results are retained, and snippets are extracted. An optional Natural Language Inference (NLI) model can then be applied to rank snippets based on their relevance, prioritizing the most pertinent evidence.\footnote{Snippets classified as either entailment or contradiction are considered highly relevant.}

\paragraph{Extensibility}
\loki's modular design enables high extensibility. Each component can be individually optimized or replaced with improved solutions as needed. For LLM-based functions, \loki supports multiple LLM APIs as well as locally deployed models. Additionally, if users require only specific functionalities, each component can be imported and used independently.

\subsection{Parallelism and Efficiency}
\loki uses parallelism to efficiently handle multiple verification requests at the same time. By using Python's \texttt{asyncio} library, the system can make asynchronous API calls, enabling concurrent processing without blocking the main thread. The LLM call function manages these calls and ensures compliance with rate limits with a user-defined maximum request per minute and request window. A traffic queue keeps track of active requests and removes old entries when they exceed the time window. This approach allows \loki to process high volumes of fact-checking requests efficiently, improving overall system performance and scalability.

We also parallelize all independent components and functions, as shown in \figref{fig:parallelism}. This modular design allows different parts of the fact-checking process, such as claim decomposition, query generation, and evidence retrieval, to be executed concurrently. Ideally, the entire fact-checking process can be completed within the time required for three LLM calls and one web query, significantly reducing the overall response time.

\subsection{Multilingual Support}
\loki supports multilingual scenarios. Developers can easily adapt the system by modifying prompts for the target language and performing minimal testing. We provide a comprehensive development guide to facilitate this process. Currently, \loki supports both English and Chinese, with the potential for easy adaptation to other languages.

\subsection{Human-in-the-Loop Design}

One of the key features of \loki is its human-in-the-loop design, which prioritizes transparency and user engagement to enhance human decision-making rather than replace it.

Unlike fully automated solutions that aim to deliver a final verdict without intermediate steps, \loki presents critical information and insights at each step, to assist users in making well-informed decisions. This approach is especially beneficial for users such as journalists and content moderators, who need reliable tools to verify information while retaining control over the final judgment.

\begin{figure*}[t]
    \centering
    \includegraphics[width=0.9\textwidth]{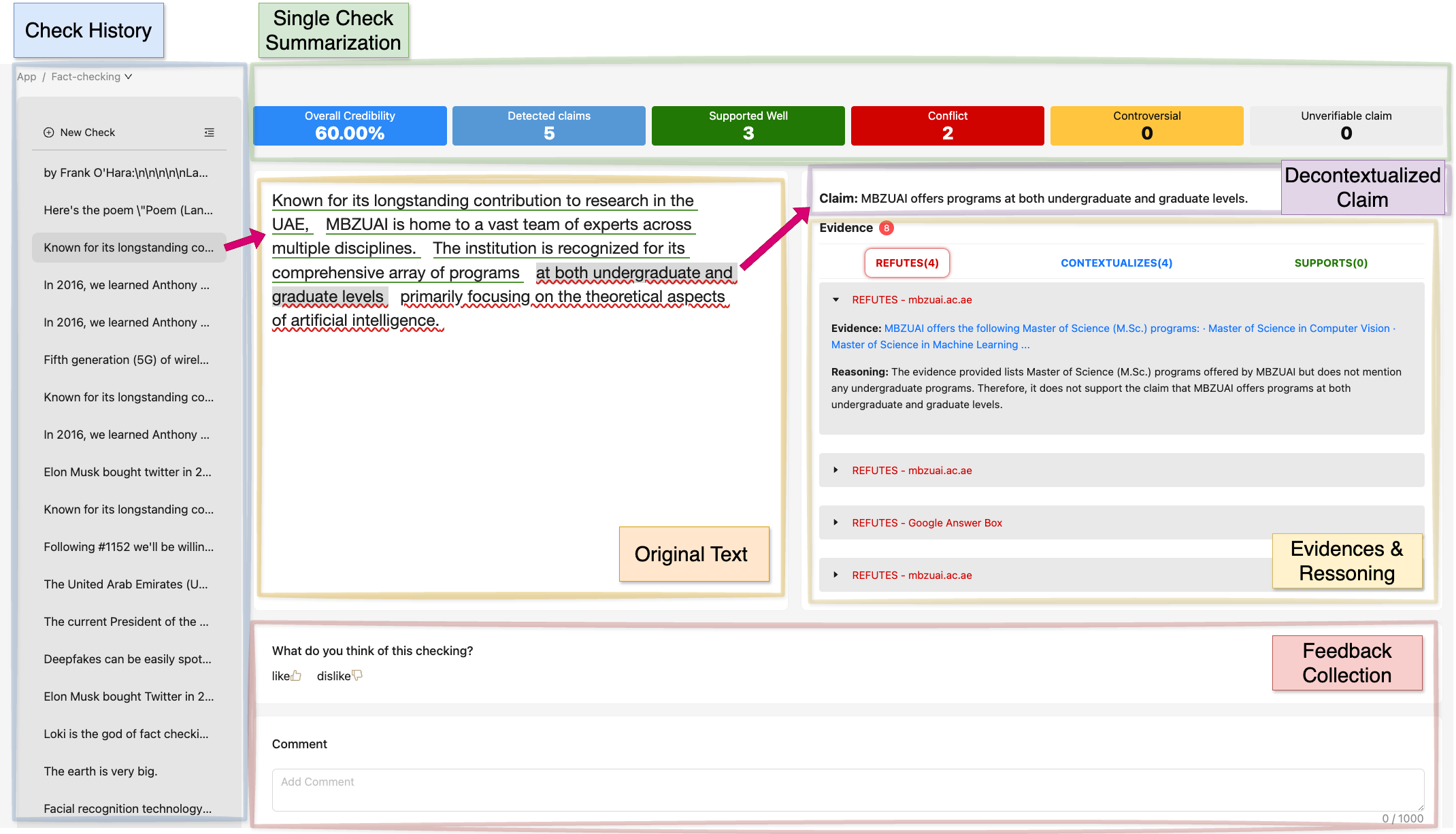}
    \caption{User interface of \loki. The interface includes sections for Check History (left panel) showing past checks, and Single Check Summarization (top center) summarizing key metrics such as overall credibility and detected claims. Original Text (left center), and the Decontextualized Claim (top right) which isolates the specific claim being evaluated. The Evidence \& Reasoning panel (right) provides detailed evidence that refutes, contextualizes, or supports the claim, sourced from various documents, while the Feedback Collection section (bottom) allows users to provide feedback.}
    \label{fig:loki_ui}
\end{figure*}

To facilitate user understanding and engagement, \loki offers information across four distinct levels:

\textbf{Level 1 --- Overall Credibility Score:} At the highest level, \loki provides a single percentage score representing the overall credibility of the input text, based on the proportion of claims verified as \emph{well-supported}, \emph{conflicting}, or \emph{controversial}, offering a quick summary of the text’s reliability.

\textbf{Level 2 --- Claim-Level Analysis:} Claims are classified and displayed numerically beside the overall score, with visual cues highlighting each claim directly in the original text, allowing users to quickly identify areas requiring further scrutiny.

\textbf{Level 3 --- Evidence-Level Insight:} For each claim, \loki presents supporting or refuting evidence from various sources. It also provides contextual information to help users understand the broader background, enabling them to make informed judgments.

\textbf{Level 4 --- Detailed Evidence Breakdown:} At the finest granularity, \loki presents detailed information for each piece of evidence, including the source, relevant paragraphs, and the rationale for why the evidence supports or contradicts the claim.

Our interaction design is guided by three core principles: (1) Transparency and Trust --- \loki provides a clear breakdown of information at multiple levels, allowing users to trace the decision-making process and build trust in the system's outputs. (2) Hierarchical Presentation --- \loki presents information in layers to avoid overwhelming users, enabling them to ``zoom in and out'' to different levels of detail for a seamless experience. (3) Assisting, not Replacing: \loki does not aim to make judgments for the users, but rather to assist them with organized and structured information.

\section{System Evaluation and Applications}

\begin{table*}[t]
\centering
\resizebox{\textwidth}{!}{%
\small

\begin{tabular}{lcccccc@{\hspace{6pt}}cccc@{\hspace{10pt}}cccc@{\hspace{6pt}}ccc}
\toprule
\multirow{3}{*}{\textbf{Framework}} & \multirow{3}{*}{\textbf{Verifier}} & \multirow{3}{*}{\textbf{\begin{tabular}[c]{@{}c@{}}Source/\\ Retriever\end{tabular}}} & \multicolumn{7}{c}{\textbf{\factcheckbench}} & & \multicolumn{7}{c}{\textbf{\factoolqa}} \\
 &  &  & \multicolumn{3}{c}{\textbf{Label = True}} & & \multicolumn{3}{c}{\textbf{Label = False}} & & \multicolumn{3}{c}{\textbf{Label = True}} & & \multicolumn{3}{c}{\textbf{Label = False}}  \\
 &  &  & P & R & F1 & & P & R & F1 & & P & R & F1 & & P & R & F1 \\ \midrule
Random & -- & -- & 0.79 & 0.43 & 0.56 & & 0.18 & 0.52 & 0.27 & & 0.79 & 0.56 & 0.66 & & 0.28 & 0.54 & 0.37  \\
Always True & -- & -- & 0.81 & 1.00 & 0.88 & & 0.00 & 0.00 & 0.00 & & 0.76 & 1.00 & 0.86 & & 0.00 & 0.00 & 0.00  \\
Always False & -- & -- & 0.00 & 0.00 & 0.00 & & 0.19 & 1.00 & 0.33 & & 0.00 & 0.00 & 0.00 & & 0.24 & 1.00 & 0.39  \\ \midrule
\factscore & \llama3-Inst 8B & Wiki/BM25 & 0.87 & 0.74 & 0.80 & & 0.34 & 0.56 & 0.42 & & 0.82 & 0.68 & 0.74 & & 0.34 & 0.52 & 0.41  \\
\factool & \llama3-Inst 8B & Web/Serper & 0.88 & 0.80 & \textbf{0.84} & & 0.40 & 0.56 & 0.47 & & \textbf{0.93} & 0.38 & 0.54 & & 0.32 & \textbf{0.91} & 0.47  \\
\factscore & GPT-3.5-Turbo & Wiki/BM25 & 0.87 & 0.67 & 0.76 & & 0.31 & 0.60 & 0.41 & & 0.82 & 0.58 & 0.68 & & 0.31 & 0.59 & 0.40  \\
\factool & GPT-3.5-Turbo & Web/Serper & 0.89 & 0.74 & 0.81 & & 0.37 & 0.62 & 0.46 & & 0.92 & 0.59 & 0.72 & & 0.39 & 0.84 & 0.53  \\
\factcheckgpt & \gptfour & Web/SerpAPI & 0.90 & 0.71 & 0.79 & & 0.52 & 0.80 & \textbf{0.63} & & 0.88 & \textbf{0.88} & \textbf{0.88} & & \textbf{0.63} & 0.63 & \textbf{0.63}  \\
\factool & \gptfouro & Web/Serper & \textbf{0.91} & 0.58 & 0.71 & & 0.47 & \textbf{0.87} & 0.61 & & 0.9 & 0.69 & 0.78 & & 0.44 & 0.77 & 0.56  \\ \midrule
\perplexityai & Sonar-online & Web & \textbf{0.93} & 0.73 & 0.83 & & 0.40 & 0.76 & 0.53 & & 0.82 & \textbf{0.88} & 0.85 & & 0.50 & 0.38 & 0.43  \\ \midrule
\loki & \gptfouro & Web/Serper & 0.84 & \textbf{0.83} & \textbf{0.84} & & \textbf{0.63} & 0.64 & \textbf{0.63} & & 0.89 & 0.80 & 0.85 & & 0.53 & 0.70 & 0.60  \\
\bottomrule
\end{tabular}

}
\caption{
The performance of fact-checkers for human-annotated claims in \factcheckbench, \factoolqa, and \felmwk, judging whether or not a claim is factually true or false with external knowledge (Wikipedia or Web articles) as evidence. 
\gptfour refers to \texttt{gpt-4-turbo-2024-04-09}, \gptfouro refers to \texttt{gpt-4o-2024-05-13}.
}
\label{tab:eval_perf}
\end{table*}

\begin{table}[t!]
\centering
\resizebox{1\columnwidth}{!}{
\begin{tabular}{llccc}
\toprule
\textbf{Dataset $\downarrow$}  & \textbf{Domain} & \textbf{\#True} & \textbf{\#False} & \textbf{Total} \\
\midrule
\factoolqa & History, geography, biology, science & 177 & 56 & 233 \\
\factcheckbench & Technology, history, science, sports & 472 & 206 & 678 \\
\bottomrule
\end{tabular}
}
\caption{The number of claim labels as well as the involved domain in \factoolqa, \felmwk and \factcheckbench.}
\label{tab:factbench-statistics}
\end{table}

\begin{table}[t!]
\centering
\resizebox{\columnwidth}{!}{
\begin{tabular}{lcccc}
\toprule
\textbf{Fact-Checker $\downarrow$} & \textbf{\begin{tabular}[c]{@{}c@{}}Web search\\  (\# queries)\end{tabular}} & \textbf{\begin{tabular}[c]{@{}c@{}}\# Prompt\\  Tokens\end{tabular}} & \textbf{\begin{tabular}[c]{@{}c@{}}\# Completion\\  Tokens\end{tabular}} & \textbf{\begin{tabular}[c]{@{}c@{}}Time \\ (s / sample)\end{tabular}} \\ \midrule
\factool & 3.5 $\pm$ 2.3 & 1895 $\pm$ 911 & 283 $\pm$ 212 & 12.38 $\pm$ 46.84 \\
\loki & 4.2 $\pm$ 2.6 & 6518 $\pm$ 4055 & 818 $\pm$ 565 & 8.32 $\pm$ 8.17 \\ \bottomrule
\end{tabular}%
}
\caption{Comparison between \factool vs. \loki (both are using \gptfouro and Serper for verification and retrieval) in terms of cost and latency. Metrics are presented in the form of their average and standard deviation ($\mu\pm\sigma$).}
\label{tab:latency_cost}
\end{table}

We performed a comprehensive evaluation for \loki across several key dimensions, including fact-checking accuracy (measured by precision, recall and F1-score), processing latency (execution wall time), and potential operation cost (number of tokens and queries consumed by API invocation).

\paragraph{Dataset}
Specifically, our experiments are carried out using two well-known automatic fact-checking evaluation datasets (\factoolqa~\citep{chern2023factool} and \factcheckbench~\citep{wang2023factcheck}), each covering a broad range of domains (see \Cref{tab:factbench-statistics} for details). In addition to the prompts and LLM responses, these datasets also provide human-annotated claims for each response, along with their labels (\texttt{True} or \texttt{False}), enabling a more granular evaluation at the claim level rather than at the response level.

\paragraph{Baseline Systems}

We selected three fact-checking systems and a commercial retrieval-augmented generative model (Perplexity.ai) as baseline systems. While the three fact-checking systems share a similar pipeline, they differ in their components at various stages compared to \loki, such as the LLM used in the verifier, data sources, and the retriever API being used. To ensure a comprehensive evaluation, we assess the baseline systems across multiple configurations of these components.

\paragraph{Evaluation Protocol}
To guarantee a fair evaluation across all fact-checking systems, we standardized the evaluation process by bypassing the step of extracting atomic claims from documents. As a result, all systems receive a claim as input and are expected to predict whether the claim is factual. Note that for frameworks like \loki that predict a factuality score (ranging from 0 to 1), we apply a threshold of 0.8 to convert the score into a binary label.

All systems were evaluated on the same computing node of the M3 Cluster,\footnote{https://massive.org.au/} with the following configuration: 2 $\times$ Intel Xeon Gold 6330 CPUs, 200 GB of memory, and 2 A100 80GB GPUs.

\paragraph{Fact-checking Performance}
Experimental results for the evaluation of fact-checking accuracy on different systems are presented in \Cref{tab:eval_perf}. Although \loki does not outperform all frameworks on every metric, its overall performance is comparable to the SOTA open-source baseline \factcheckgpt. It's important to note that Factcheck-GPT uses the more advanced SerpAPI, which retrieves higher-quality documents, significantly improving its performance. However, the cost of SerpAPI (\$0.015 per search) is much higher than Serper (\$0.001 per search), making \loki the more economical option in terms of cost-effectiveness. When compared to \factool, which uses the same configuration, \loki outperforms it in most of the metrics across three datasets, demonstrating the effectiveness of our design.

\subsection{Cost and Latency}

In \Cref{tab:latency_cost}, we provide a detailed comparison of the number of queries, token consumption during LLM API calls, and the average processing time per claim for both \loki and \factool across the two datasets. We did not include comparisons with other frameworks because only \factool and \ofc support asynchronous processing, which results in significantly lower processing times than the other frameworks.\footnote{According to \citet{wang2024openfactcheck}, a cascade pipeline like \factcheckgpt is normally $15.7\times$ slower than an asynchronous pipeline.} Since \ofc's fact-checking pipeline is largely derived from other frameworks, including \factool, we consider \factool sufficiently representative for comparison.

The results show that \loki consumes more queries and more tokens than \factool, primarily because \loki retrieves and verifies a greater number of queries and evidence. This also explains why \loki achieves superior fact-checking performance. Despite handling larger volumes of data, \loki's efficient implementation ensures that its average processing time is substantially lower than that of \factool, highlighting its better design, with the asynchronous processing pipeline.



\section{Conclusion}

In this paper, we present \loki, a novel open-source Python tool that offers a human-centered approach to fact-checking. \loki integrates a semi-automated five-step pipeline that balances efficiency with the need for human judgment, particularly for general users like journalists and content moderators. Experimental results show that \loki achieves competitive performance against state-of-the-art systems, offering an optimized and adaptable solution for tackling misinformation. As an open-source tool licensed under MIT, \loki is accessible for further development, aiming to enhance fact-checking capabilities and support informed decision-making continuously.

\bibliography{custom}

\begin{thebibliography}{14}
\providecommand{\natexlab}[1]{#1}

\bibitem[{Augenstein et~al.(2023)Augenstein, Baldwin, Cha, Chakraborty, Ciampaglia, Corney, DiResta, Ferrara, Hale, Halevy, Hovy, Ji, Menczer, Miguez, Nakov, Scheufele, Sharma, and Zagni}]{augenstein2023factualitychallengeseralarge}
Isabelle Augenstein, Timothy Baldwin, Meeyoung Cha, Tanmoy Chakraborty, Giovanni~Luca Ciampaglia, David Corney, Renee DiResta, Emilio Ferrara, Scott Hale, Alon Halevy, Eduard Hovy, Heng Ji, Filippo Menczer, Ruben Miguez, Preslav Nakov, Dietram Scheufele, Shivam Sharma, and Giovanni Zagni. 2023.
\newblock \href {https://arxiv.org/abs/2310.05189} {Factuality challenges in the era of large language models}.
\newblock \emph{Preprint}, arXiv:2310.05189.

\bibitem[{Chern et~al.(2023)Chern, Chern, Chen, Yuan, Feng, Zhou, He, Neubig, and Liu}]{chern2023factool}
I{-}Chun Chern, Steffi Chern, Shiqi Chen, Weizhe Yuan, Kehua Feng, Chunting Zhou, Junxian He, Graham Neubig, and Pengfei Liu. 2023.
\newblock \href {https://doi.org/10.48550/arXiv.2307.13528} {Factool: Factuality detection in generative {AI} - {A} tool augmented framework for multi-task and multi-domain scenarios}.
\newblock \emph{CoRR}, abs/2307.13528.

\bibitem[{Dhuliawala et~al.(2023)Dhuliawala, Komeili, Xu, Raileanu, Li, Celikyilmaz, and Weston}]{dhuliawala2023chain}
Shehzaad Dhuliawala, Mojtaba Komeili, Jing Xu, Roberta Raileanu, Xian Li, Asli Celikyilmaz, and Jason Weston. 2023.
\newblock \href {https://arxiv.org/abs/2309.11495} {Chain-of-verification reduces hallucination in large language models}.
\newblock \emph{arXiv preprint arXiv:2309.11495}.

\bibitem[{Fadeeva et~al.(2024)Fadeeva, Rubashevskii, Shelmanov, Petrakov, Li, Mubarak, Tsymbalov, Kuzmin, Panchenko, Baldwin, Nakov, and Panov}]{fadeeva-etal-2024-fact}
Ekaterina Fadeeva, Aleksandr Rubashevskii, Artem Shelmanov, Sergey Petrakov, Haonan Li, Hamdy Mubarak, Evgenii Tsymbalov, Gleb Kuzmin, Alexander Panchenko, Timothy Baldwin, Preslav Nakov, and Maxim Panov. 2024.
\newblock \href {https://doi.org/10.18653/v1/2024.findings-acl.558} {Fact-checking the output of large language models via token-level uncertainty quantification}.
\newblock In \emph{Findings of the Association for Computational Linguistics ACL 2024}, pages 9367--9385, Bangkok, Thailand and virtual meeting. Association for Computational Linguistics.

\bibitem[{Gao et~al.(2022)Gao, Dai, Pasupat, Chen, Chaganty, Fan, Zhao, Lao, Lee, Juan et~al.}]{gao2022attributed}
Luyu Gao, Zhuyun Dai, Panupong Pasupat, Anthony Chen, Arun~Tejasvi Chaganty, Yicheng Fan, Vincent~Y Zhao, Ni~Lao, Hongrae Lee, Da-Cheng Juan, et~al. 2022.
\newblock \href {https://arxiv.org/abs/2210.08726} {Attributed text generation via post-hoc research and revision}.
\newblock \emph{arXiv preprint arXiv:2210.08726}.

\bibitem[{Min et~al.(2023)Min, Krishna, Lyu, Lewis, Yih, Koh, Iyyer, Zettlemoyer, and Hajishirzi}]{min2023factscore}
Sewon Min, Kalpesh Krishna, Xinxi Lyu, Mike Lewis, Wen{-}tau Yih, Pang~Wei Koh, Mohit Iyyer, Luke Zettlemoyer, and Hannaneh Hajishirzi. 2023.
\newblock \href {https://doi.org/10.48550/arXiv.2305.14251} {Factscore: Fine-grained atomic evaluation of factual precision in long form text generation}.
\newblock \emph{CoRR}, abs/2305.14251.

\bibitem[{OpenAI(2024)}]{openai2024gpt4technicalreport}
OpenAI. 2024.
\newblock \href {https://arxiv.org/abs/2303.08774} {Gpt-4 technical report}.
\newblock \emph{Preprint}, arXiv:2303.08774.

\bibitem[{Pan et~al.(2023)Pan, Pan, Chen, Nakov, Kan, and Wang}]{pan2023riskmisinformationpollutionlarge}
Yikang Pan, Liangming Pan, Wenhu Chen, Preslav Nakov, Min-Yen Kan, and William~Yang Wang. 2023.
\newblock \href {https://arxiv.org/abs/2305.13661} {On the risk of misinformation pollution with large language models}.
\newblock \emph{Preprint}, arXiv:2305.13661.

\bibitem[{Wang et~al.(2024{\natexlab{a}})Wang, Reddy, Mujahid, Arora, Rubashevskii, Geng, Afzal, Pan, Borenstein, Pillai, Augenstein, Gurevych, and Nakov}]{wang2024factcheckbenchfinegrainedevaluationbenchmark}
Yuxia Wang, Revanth~Gangi Reddy, Zain~Muhammad Mujahid, Arnav Arora, Aleksandr Rubashevskii, Jiahui Geng, Osama~Mohammed Afzal, Liangming Pan, Nadav Borenstein, Aditya Pillai, Isabelle Augenstein, Iryna Gurevych, and Preslav Nakov. 2024{\natexlab{a}}.
\newblock \href {https://arxiv.org/abs/2311.09000} {Factcheck-bench: Fine-grained evaluation benchmark for automatic fact-checkers}.
\newblock \emph{Preprint}, arXiv:2311.09000.

\bibitem[{Wang et~al.(2023)Wang, Reddy, Mujahid, Arora, Rubashevskii, Geng, Afzal, Pan, Borenstein, Pillai et~al.}]{wang2023factcheck}
Yuxia Wang, Revanth~Gangi Reddy, Zain~Muhammad Mujahid, Arnav Arora, Aleksandr Rubashevskii, Jiahui Geng, Osama~Mohammed Afzal, Liangming Pan, Nadav Borenstein, Aditya Pillai, et~al. 2023.
\newblock Factcheck-gpt: End-to-end fine-grained document-level fact-checking and correction of llm output.
\newblock \emph{arXiv preprint arXiv:2311.09000}.

\bibitem[{Wang et~al.(2024{\natexlab{b}})Wang, Wang, Iqbal, Georgiev, Geng, and Nakov}]{wang2024openfactcheck}
Yuxia Wang, Minghan Wang, Hasan Iqbal, Georgi Georgiev, Jiahui Geng, and Preslav Nakov. 2024{\natexlab{b}}.
\newblock \href {https://doi.org/10.48550/ARXIV.2405.05583} {Openfactcheck: {A} unified framework for factuality evaluation of llms}.
\newblock \emph{CoRR}, abs/2405.05583.

\bibitem[{Wang et~al.(2024{\natexlab{c}})Wang, Wang, Manzoor, Georgiev, Das, and Nakov}]{wang2024factuality-survey}
Yuxia Wang, Minghan Wang, Muhammad~Arslan Manzoor, Georgi Georgiev, Rocktim~Jyoti Das, and Preslav Nakov. 2024{\natexlab{c}}.
\newblock \href {https://doi.org/10.48550/ARXIV.2402.02420} {Factuality of large language models in the year 2024}.
\newblock \emph{CoRR}, abs/2402.02420.

\bibitem[{Wei et~al.(2024{\natexlab{a}})Wei, Yang, Song, Lu, Hu, Tran, Peng, Liu, Huang, Du, and Le}]{wei2024longformfactuality}
Jerry Wei, Chengrun Yang, Xinying Song, Yifeng Lu, Nathan Hu, Dustin Tran, Daiyi Peng, Ruibo Liu, Da~Huang, Cosmo Du, and Quoc~V. Le. 2024{\natexlab{a}}.
\newblock \href {https://doi.org/10.48550/ARXIV.2403.18802} {Long-form factuality in large language models}.
\newblock \emph{CoRR}, abs/2403.18802.

\bibitem[{Wei et~al.(2024{\natexlab{b}})Wei, Yang, Song, Lu, Hu, Tran, Peng, Liu, Huang, Du, and Le}]{wei2024longformsafe}
Jerry Wei, Chengrun Yang, Xinying Song, Yifeng Lu, Nathan Hu, Dustin Tran, Daiyi Peng, Ruibo Liu, Da~Huang, Cosmo Du, and Quoc~V. Le. 2024{\natexlab{b}}.
\newblock \href {https://doi.org/10.48550/ARXIV.2403.18802} {Long-form factuality in large language models}.
\newblock \emph{CoRR}, abs/2403.18802.

\end{thebibliography}

\appendix
\newpage

\section{Usage of \loki}

\paragraph{Used as a Library} \loki can be integrated as a Python library, allowing developers to incorporate its functionalities directly into their applications. \figref{fig:lib} provides an example of how to use FactCheck as a library for verifying text within a Python script.

\begin{figure}[H]
\begin{minted}[
frame=lines,
framesep=2mm,
numbersep=3pt,
linenos,
baselinestretch=1.1,
bgcolor=bg,
fontsize=\scriptsize,
breaklines,
]{python}
from factcheck import FactCheck

factcheck_instance = FactCheck()

# Example text
text = "Your text here"

# Run the fact-check pipeline
results = factcheck_instance.check_response(text)
print(results)
\end{minted}
\caption{Using FactCheck as a Library}
\label{fig:lib}
\end{figure}

\paragraph{Used as a Web App} \loki can be deployed as a web application, providing a user-friendly interface for interacting with the tool. \figref{fig:web_app} illustrates the command used to start the web application. The UI of open-source \loki is introduced in \appref{sec:open_ui}

\begin{figure}[H]
\begin{minted}[
frame=lines,
framesep=2mm,
numbersep=3pt,
linenos,
baselinestretch=1.1,
bgcolor=bg,
fontsize=\scriptsize,
breaklines,
]{bash}
python webapp.py --api_config demo_data/api_config.yaml
\end{minted}
\caption{Running FactCheck as a Web App}
\label{fig:web_app}
\end{figure}

\begin{figure*}[t]
    \centering
    \includegraphics[width=1\textwidth]{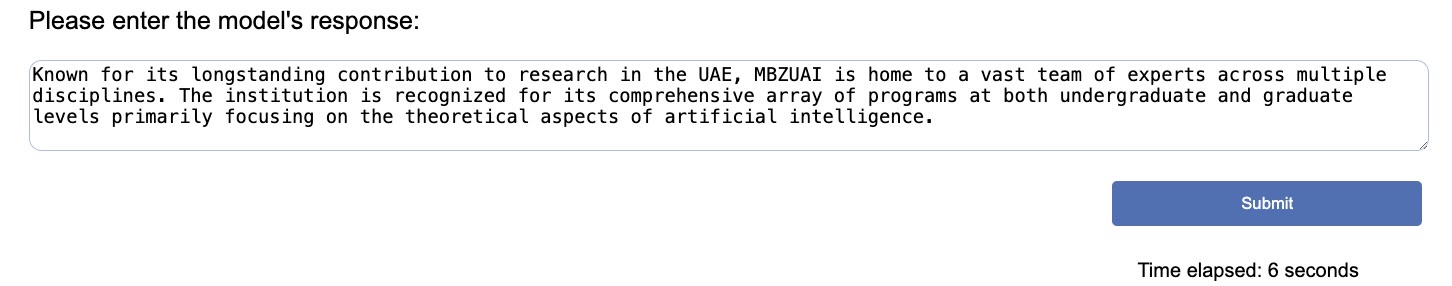}
    \caption{The user interface of \loki, submission page.}
    \label{fig:loki_ui1}
\end{figure*}

\begin{figure*}[t]
    \centering
    \includegraphics[width=1\textwidth]{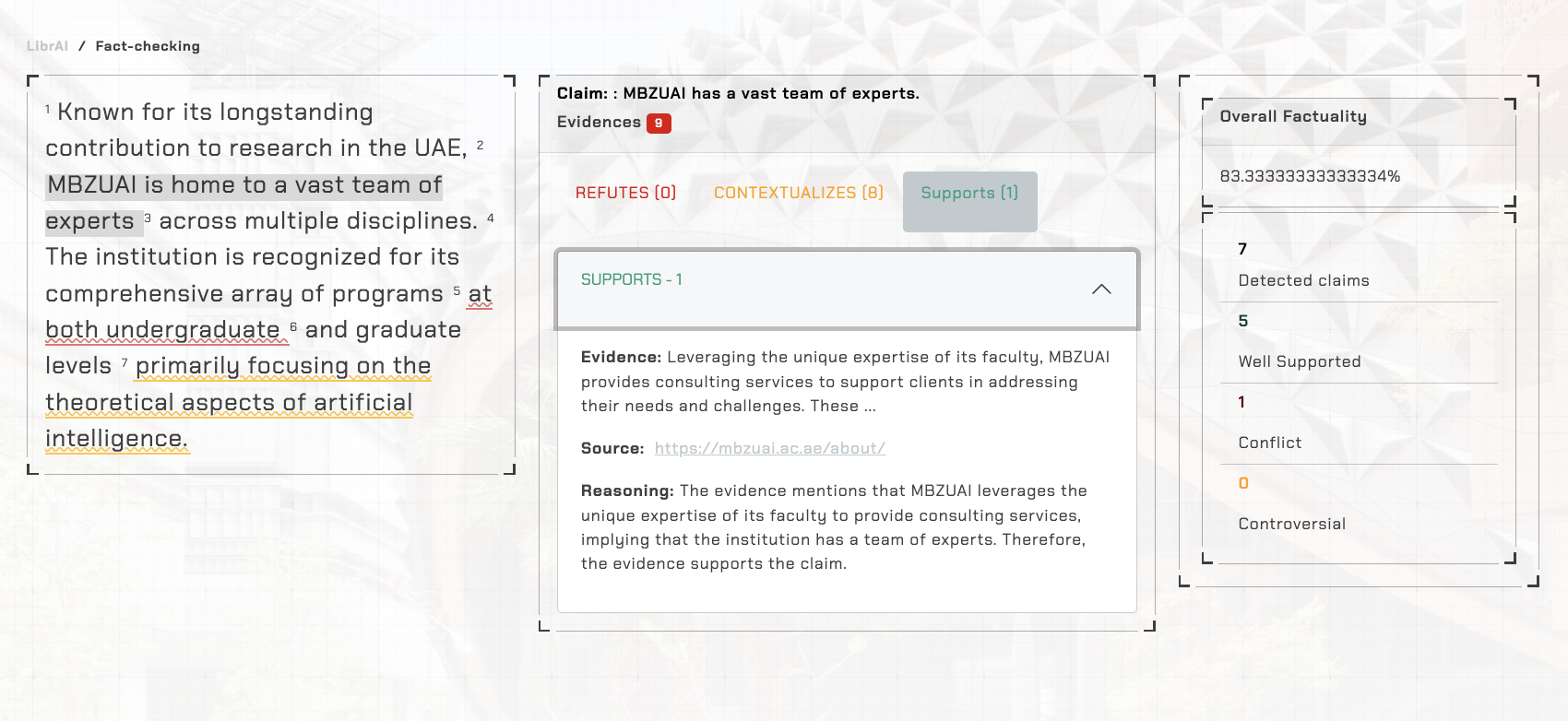}
    \caption{The user interface of \loki, result page.}
    \label{fig:loki_ui2}
\end{figure*}

\paragraph{Multimodal Usage} \loki supports multimodal input, enabling it to process and verify information from various sources, including text, speech, images, and videos. The system can be invoked using different modes specified by the \mintinline{bash}{--modal} argument. \figref{fig:multimodal} demonstrates the usage of FactCheck across these different input modalities

\begin{figure}[H]
\begin{minted}[
frame=lines,
framesep=2mm,
numbersep=3pt,
linenos,
baselinestretch=1.1,
bgcolor=bg,
fontsize=\scriptsize,
breaklines,
]{bash}
# String
python -m factcheck --modal string --input "MBZUAI is the first AI university in the world"
# Text
python -m factcheck --modal text --input demo_data/text.txt
# Speech
python -m factcheck --modal speech --input demo_data/speech.mp3
# Image
python -m factcheck --modal image --input demo_data/image.webp
# Video
python -m factcheck --modal video --input demo_data/video.m4v
\end{minted}
\caption{Multimodal Usage of FactCheck}
\label{fig:multimodal}
\end{figure}

\section{User Interface of \loki}\label{sec:open_ui}

\figref{fig:loki_ui1} and \figref{fig:loki_ui2} show the user interface of \loki (open-source version), which consists of two main pages: the submission page and the result page. The submission page allows users to input text for fact-checking, while the result page displays the decomposed claims, the evidence retrieved for each claim, and the overall credibility score of the text. The user interface is designed to be user-friendly, providing a clear and intuitive experience for users to interact with the system.

\end{document}